\documentclass[conference]{IEEEtran}
\IEEEoverridecommandlockouts
\usepackage{cite}
\usepackage{amsmath,amssymb,amsfonts}
\usepackage{algorithmic}
\usepackage{listings}
\usepackage{graphicx}
\usepackage{textcomp}
\usepackage{xcolor}
\usepackage{amsmath}
\usepackage{amssymb}
\usepackage{booktabs}
\usepackage{todonotes}
\usepackage[small]{caption}
\usepackage{float}
\usepackage{bm}
\usepackage{subcaption}
\usepackage{hyperref}
\usepackage{todonotes}

\def\BibTeX{{\rm B\kern-.05em{\sc i\kern-.025em b}\kern-.08em
    T\kern-.1667em\lower.7ex\hbox{E}\kern-.125emX}}

\newfloat{lstfloat}{htbp}{lop}
\floatname{lstfloat}{Algorithm}
\makeatletter
\let\@oldmaketitle\@maketitle% Store \@maketitle
\renewcommand{\@maketitle}{\@oldmaketitle% Update \@maketitle to insert...
\vspace{-0.78cm} % Might have to change this in the future.
\setcounter{figure}{0}
\begin{center}
    \includegraphics[width=2\columnwidth]{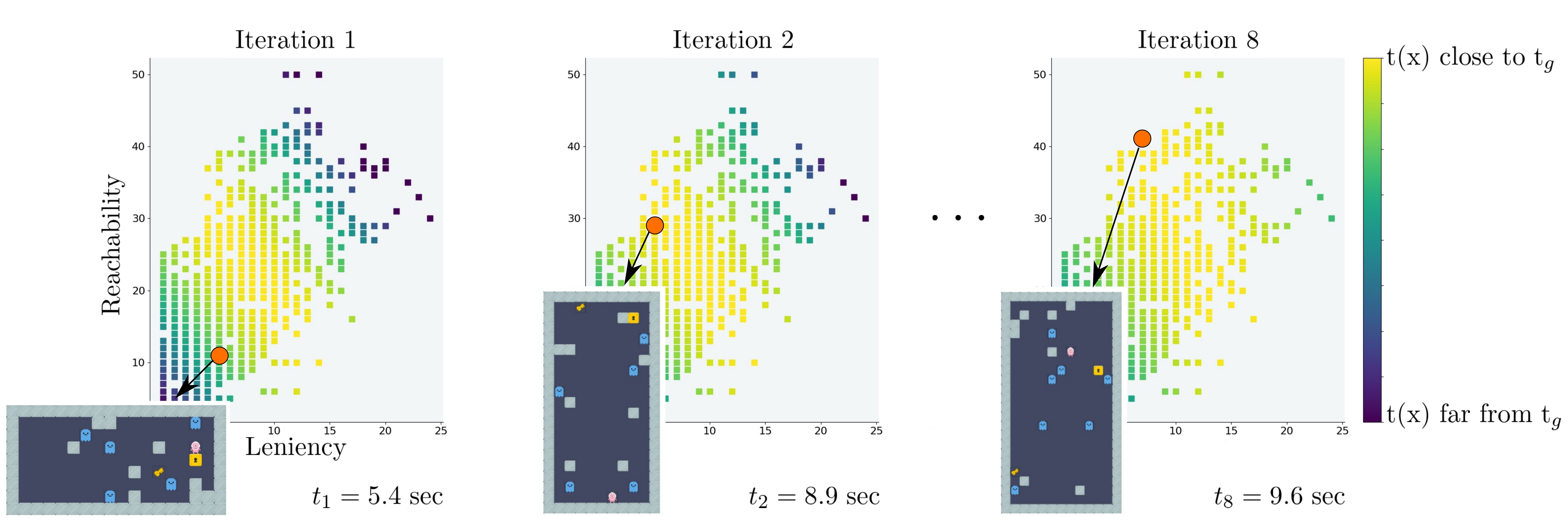} %sweet spot for 
    \captionof{figure}{\textbf{Fast Bayesian Content Adaption (FBCA).} Our Bayesian Optimization approach can adapt the level of a simple Roguelike game to a user, such that it takes the player approximately $t_g=10$ seconds to solve. Our approach models the player's completion time $t(x)$ using Gaussian Process Regression and a modified acquisition function in a Bayesian Optimization scheme, starting with a simple prior. Shown here is the session of a single player: the acquisition function first suggests a level with 5 enemies (leniency) which an A* agent can solve in 11 steps (reachability). This level takes the player roughly 5 seconds to solve. We update the prior with this information and query the acquisition function for the next level to show. After 8 levels, our system finds a level that takes this player 9.6 seconds to solve.}\label{fig:banner}
\end{center}
}
\makeatother

\begin{document}

% An idea: it should be something like Fast and Bespoke Game content adaption through Bayesian Optimization.
\title{Fast Game Content Adaptation Through Bayesian-based Player Modelling
}
 
\author{\IEEEauthorblockN{Miguel González-Duque}
\IEEEauthorblockA{\textit{Creative AI Lab} \\
\textit{IT University of Copenhagen}\\
Copenhagen, Denmark \\
migd@itu.dk}
\and
\IEEEauthorblockN{Rasmus Berg Palm}
\IEEEauthorblockA{\textit{Creative AI Lab} \\
\textit{IT University of Copenhagen}\\
Copenhagen, Denmark \\
rasmb@itu.dk}
\and
\IEEEauthorblockN{Sebastian Risi}
\IEEEauthorblockA{\textit{Creative AI Lab} \\
\textit{IT University of Copenhagen}\\
Copenhagen, Denmark \\
sebr@itu.dk}
}

\maketitle

\begin{abstract}
In games, as well as many user-facing systems, adapting content to users' preferences and experience is an important challenge. This paper explores a novel method to realize this goal in the context of dynamic difficulty adjustment (DDA). Here the aim is to constantly adapt the content of a game to the skill level of the player, keeping them engaged by avoiding states that are either too difficult or too easy. Current systems for DDA rely on expensive data mining, or on hand-crafted rules designed for particular domains, and usually adapts to keep players \textit{in the flow}, leaving no room for the designer to present content that is purposefully easy or difficult. This paper presents Fast Bayesian Content Adaption (FBCA), a 
%Bayesian Optimization-based 
system for DDA that is agnostic to the domain and that can target particular difficulties. We deploy this framework in two different domains: the puzzle game Sudoku, and a simple Roguelike game. By modifying the acquisition function's optimization, we are reliably able to present a content with a bespoke difficulty for players with different skill levels in less than five iterations for Sudoku and fifteen iterations for the simple Roguelike. Our method significantly outperforms simpler DDA heuristics  with the added benefit of maintaining a model of the user. These results point towards a promising alternative for content adaption in a variety of different domains.
\end{abstract}


\section{Introduction}

% TODO: Clearly state the problem being solved.
The problem of creating interactive media that adapts to the user has several applications, ranging from increasing the engagement of visitors of web applications to creating tailored experiences for students in academic settings \cite{Pastushenko2019}. One of these applications is Dynamic Difficulty Adjustment (DDA) \cite{Hunicke2004}, which consists of adapting the contents of a video game to match the skill level of the player. If the game presents tasks that are too difficult or too easy, it might risk losing the player due to frustration or boredom.

% \todo{what do you mean they don't generalise? you can apply MCTS to many different domains}
Current approaches to DDA focus on specific domains (e.g. MOBAs \cite{SILVA2017103}, Role-Playing games \cite{Zook2012data} or fighting games \cite{Demediuk2017}), and use agents and techniques that either do not generalize (such as planning agents requiring forward models \cite{Hao2010PacMan}), or rely on an expensive process of gathering data from players \textit{before} the optimization can take place \cite{Jennings-Teats2010,moon2020dynamic}. Also relevant is the fact that most of these approaches focus on maximized engagement and achieving flow states. However, sometimes 
%leaving less room for games in which 
the designer's intent might be to purposefully present content that is difficult (i.e.\  out-of-flow) for a particular player \cite{AnthropyandClark2014}.

Bayesian Optimization has been recently proposed as a promising approach to DDA, since it does not rely on previously gathered information on either user or domain, and can be deployed online with only minimal specifications about the game in question \cite{gonzalez2020finding}. However, so far this approach has only been tested with AI agents and in a single domain, while only allowing for one possible difficulty target.

We propose a new Bayesian-based method for Fast Bayesian Content Adaption and test it on DDA with human players. The method maintains a simple model of the player and leverages it for optimizing content towards a target difficulty on-the-fly in a \textit{domain-agnostic} fashion. Fig.~\ref{fig:banner} illustrates how the proposed approach works: (1) Players are presented with levels that are predicted to have the right difficulty by the underlying probabilistic model; (2) the model is updated once new data about the player's performance arrives; (3) steps 1--2 are repeated until a level with the desired target difficulty is found.  
%an underlying probabilistic model predicts may have the right difficulty, and the model is updated once new data about the player's performance arrives.

We test this novel approach on two domains: the puzzle game Sudoku, and levels for a simple Roguelike game. Our results
%albeit noisy, 
show that Bayesian Optimization is a promising alternative for domain-agnostic automatic difficulty adjustment.

% In other words, we present a method for Fast Bayesian Content Adaption based on Bayesian Optimization, which optimizes a transformation of the modeled function (by modifying the underlying acquisition function). This change allows us to target any difficulty in an online fashion.\todo{rewrite this summary, it could be better}

\section{Methods and Related Work}\label{sec:related_work}

\subsection{Related Work on Difficulty Adjustment}

% \todo{is there any other work modelling players/users with bayesian approaches?}
Dynamic Difficulty Adjustment (DDA) consists of adapting the difficulty of a game to the skill level of the player, trying to keep a \textit{flow} state (i.e.\ a psychological state in which users solve tasks that match their ability). DDA algorithms work by \textit{predicting and intervening} \cite{Xue2017}. They model a so-called \textit{challenge function} that stands as a proxy for difficulty (e.g. win rate, health lost, hits received, completion time) and intervene the game so as to match a particular target for this challenge function \cite{Zohaib2018}.

Hunicke \cite{Hunicke2005} points out that DDA can help players to retain a ``sense of agency and accomplishment'', presenting a system called \textit{Hamlet} 
%The case for DDA in academic research was proposed in 2004 with the advancement of \textit{Hamlet}, a system 
that leveraged inventory theory to present content to players in \textit{Half-Life} \cite{Hunicke2004,Hunicke2005}. Since then, several other approaches and methods have been presented and studied in this context, including alternating between different-performing AI agents in MOBA games \cite{SILVA2017103}, player modelling via data gathering or meta-learning techniques \cite{Jennings-Teats2010,moon2020dynamic}, and artificially restricting NPCs that are implemented as planning agents such as MCTS \cite{Hao2010PacMan,Demediuk2017}. Other DDA methods model the player using probabilities such as the 
the probabilities that a player would re-try, churn, or win a particular level in a mobile game \cite{Xue2017}; the resulting probabilistic graph is then used to maximize engagement.

Dynamic Difficulty Adjustment is a particular instance of the larger field of automatic content creation and adaption. 
%, and several works have been done in this, more general direction. 
Examples in this area include work on evolving and evaluating racetracks for different player models \cite{Togelius2007racing}.
% \todo{Maybe just say "Another example is ...", without mentioning player driven PCG}
In another example, Shaker et al. optimize the game design features of platformer games towards \textit{fun} levels, where fun is defined using player models trained on questionnaires \cite{Shaker2010}.
% Optimizing player experience has also been the focus of the player-driven Procedural Content Generation community, with experiments like e.g. optimizing game design features for platformer games using player models trained on questionnaires \cite{Shaker2010}.
Similar efforts have been made in the Experience Management community (with the goal of creating interactive storytelling games) \cite{Thue2007}. 

These approaches, however, either rely on gathering data beforehand and leveraging it to create a player model that is then used for optimization, or are domain-specific (e.g. storytellers or platformers). Bayesian Optimization serves as an alternative that is data efficient and flexible. Moreover, in contrast to AI-based approaches (like adjusting NPCs performance), Bayesian Optimization of game parameters does not rely on forward models.
% Another criticism of these AI-based approaches is that they rely on having forward models of the game for planning and adjusting content.\todo{instead of saying criticism I would reformulate to say that, in contrast to other approaches, our approaches does not rely on forward model...}

Bayesian Optimization has been used a tool for automatic playtesting and DDA. Zook et al. use Active Learning to fine-tune low level parameters based on human playtests \cite{ZookAutomaticPlaytesting2014}. Khajah et al. \cite{Khajah2016} apply Bayesian Optimization to find game parameters in \textit{Flappy Bird} and \textit{Spring Ninja} (e.g. distance between pipes and gap size) that would maximize engagement, measured as volunteered time. Our work differs in the fact that we build a model of player performance, instead of player engagement. With our system, designers have the affordance to target content that is difficult for a given player in a bespoke fashion.

To the best of our knowledge, our contribution is the first example of a Bayesian Optimization-based system that models levels of difficulty \emph{for particular players}, and dynamically presents bespoke content according to this model.

% The use of ML techniques for DDA has been explored in the form of using MCTS agents with varying degrees of compute allocation \cite{Demediuk2017,Hao2010PacMan}, alternating between a family of different-performing AI agents \cite{SILVA2017103}, and player modeling via data gathering \cite{Jennings-Teats2010} or meta-learning techniques \cite{moon2020dynamic}. Bayesian network methods have been explored for modelling user knowledge in interactive environments, but not in the context of DDA \cite{Rowe_Lester_2010}.
% \todo{expand this discussion}
%\cite{Zook2012data} [cite zook and Riedl] also propose a method for DDA through player modeling and generalized matrix factorization.
% Work using ad-hoc rules in the industry. Some work using inventory theory. Previous work modeling level transitions (EA people). Work using planning agents.
% \todo{We can copy some stuff from our previous paper here}
% Should probably mention this paper? \cite{moon2020dynamic}

% These approaches, however, are either domain-specific (e.g.\ AI agents for battle games), or rely on expensive (re)training of neural network models. Our approach, on the other hand, is domain-agnostic and does not rely on previously gathered data.\todo{fix this}

% In contrast to the approach by  \cite{gonzalez2020finding} that only evaluated a Bayesian optimization approach based on artificial agents, in this paper ... \todo{TODO}

\subsection{Bayesian Optimization (B.O.) using Gaussian Process Regression}\label{subsec:BO}

% Introduce B.O. through black-box optimization.
%% i.e. why we use B.O.
Bayesian Optimization is frequently used for optimization problems in which the objective function has no closed form, and can only be measured by expensive and noisy queries (e.g.\ optimizing hyperparameters in Machine Learning algorithms \cite{Snoek2012}, active learning \cite{Brochu2010},  finding compensatory behaviors in damaged robots \cite{Cully2015}). The problem we tackle in this paper is indeed black-box: we have no closed analytical form for the time it would take for a player to solve a given level, having them play a level is expensive time-wise, and a player's performance on a single level may vary if we serve it repeatedly.

% Bayesian optimization has two main components: a surrogate probabilistic model, and an acquisition function that indicates where to query next [cite Freitas]. For the surrogate model we use Gaussian Process Regression, and for the acquisition function we use the Upper Confidence Bound.
There are two main components in B.O. schemes \cite{BayesOptSurvey}: a surrogate probabilistic model that approximates the objective function, and an acquisition function that uses this probabilistic information to inform where to query next in order to maximize said objective function. A common selection for the underlying probabilistic model are Gaussian Processes \cite{Rasmussen2006GP}, and two frequently used acquisition functions are Expected Improvement (EI) and the Upper Confidence Bound (UCB). 
%In the following two items we will explain how Gaussian Process Regression works, and how these acquisition functions are defined.

\subsubsection{Gaussian Process Regression}

Practically speaking, a Gaussian process $\text{GP}$ defines a Normal distribution for every point-wise approximation of a function $t(x)$ using a prior $\mu_0(x)$ and a kernel function $k(x, x')$ (which governs the covariance matrix).

If we assume that the observations of the function for a set of points $\bm{x} = [x_i]_{i=1}^n$, denoted by $\bm{t} = [t(x_i)]_{i=1}^n = [t_i]_{i=1}^n$, are normally distributed with mean $\bm{\mu}_0 = [\mu_0(x_i)]_{i=1}^n$ and covariance matrix $K=[k(x_i, x_j)]_{i,j=1}^n$, we can approximate $t(x)$ at a new point $x^*$ by leveraging the fact that the Gaussian distributions are closed under marginalization \cite{Rasmussen2006GP}:
% \begin{equation}\label{eq:GP}
%     \tilde{f}\,|\,\tilde{x},\bm{x},\bm{f} \sim \mathcal{N}(\bm{\mu}_0 + \bm{k}_*^TK^{-1}\bm{f}, k(\tilde{x},\tilde{x}) -\bm{k}_*^TK^{-1}\bm{k}_*),
% \end{equation}
\begin{equation}\label{eq:GP_update}
\begin{aligned}
    t^*\,|\,x^*,\bm{x},\bm{t} \sim \mathcal{N}(&\bm{\mu}_0 + \bm{k}_*^T(K + \sigma_\text{noise})^{-1}\bm{t},\\
    &k(x^*,x^*) -\bm{k}_*^T(K + \sigma_\text{noise})^{-1}\bm{k}_*),
\end{aligned}
\end{equation}
where $\bm{k}_* = [k(x_i, x^*)]_{i=1}^n$ and $\sigma_\text{noise}\in\mathbb{R}^+$ is a hyperparameter.

% TODO: Change the language to "these are typical kernels used by everyone".

% \todo{this should be moved to the approach section. Assume people will skip the background/methods section if they know about GP and BO}

Two standard choices for kernel functions are the anisotropic radial basis function (RBF) $k_\text{RBF}(\bm{x},\bm{x}') = \exp(\bm{x}(\bm{\theta}I)\bm{x'}^T)$ and the linear (or Dot Product) kernel $k_\text{Linear}(\bm{x},\bm{x}') = \sigma_0 + \bm{x}^T\bm{x}'$. These kernels, alongside Gaussian Process Regression as described by Eq. (\ref{eq:GP_update}), are implemented in the open-source library \texttt{sklearn} \cite{scikit-learn}, which we use for this work.
% All hyperparameters involved (the vector of variances and lengthscales $\bm{\theta}$, $\sigma_0\in\mathbb{R}$ and $\sigma_{\text{noise}}$ in Eq. (\ref{eq:GP_update})) are estimated using Maximum Likelihood over the data, and in our experiments we use \texttt{sklearn}'s [cite sklearn] implementation of Gaussian Process Regression.

One important detail in our experimental setup is that, since the function that we plan to regress ($t(x)$) will always be positive empirically, we choose to model $\log(t(x))$ instead. We thus assume that $\log(t)$ (and not $t$) is normally distributed. This is a common trick for modeling positive functions.

\subsubsection{Acquisition Functions}

Let $\tilde{t}(x)$ be the objective function in a Bayesian Optimization scheme. If we model $\tilde{t}(x)$ using Gaussian Processes, we have access to point estimates and uncertainties (in the form of the updated mean and standard deviation described in Eq.~(\ref{eq:GP_update})). An acquisition function $\alpha(x)$ uses this probabilistic model to make informed guesses on which new point $x^*$ might produce the best outcome when maximizing $\tilde{t}(x)$. These functions balance exploration (querying points in parts of the domain that have not been explored yet) and exploitation (querying promising parts of the landscape, areas in which previous experiments have had high values for $\tilde{t}$).

In our experiments, we use two acquisition functions: Expected Improvement, defined as $\alpha_{\text{EI}}(x) = \mathbb{E}[\max(0, \tilde{t}(x) - \tilde{t}_\text{best})]$, where $\tilde{t}_\text{best}$ is the best performance seen so far, and the Upper Confidence Bound $\alpha_{\text{UCB}, \kappa} = \mu(x) + \kappa \sigma(x)$ where $\mu(x)$ and $\sigma(x)$ are the posterior mean and standard deviation of the Gaussian Process with which we model $\tilde{t}(x)$. The hyperparameter $\kappa$ measures the tradeoff between exploration and exploitation.

% ALGORITHM: our approach.
\begin{lstfloat}
\begin{lstlisting}
procedure FastBayesianContentAdaption($t_g$, $\mathcal{D}$, $\mu_0$, $k$, $\beta$):
    $\mathcal{X}$ = $\varnothing \subseteq \mathcal{D}$ // list of served contents $x$
    $\mathcal{T}$ = $\varnothing \subseteq \mathbb{R}$ // the log-time $\log(t)$ they took
    while True:
        // Start the GP and optimize its hyperparameters
        initialize GP$(\mu_0, k)$ and fit it with $(x, \log(t))\in\mathcal{X}\times\mathcal{T}$
        // Maximize the modified acq. function.
        $x_{\text{next}} = \max_{x\in\mathcal{D}}\{\beta^{t_g}(x)\}$
        // This task is the most likely to have time $t_g$.
        present task $x_{\text{next}}$ and record time $t$
        add $x_{\text{next}}$ to $\mathcal{X}$ and $\log(t)$ to $\mathcal{T}$
\end{lstlisting}
\caption{Pseudocode for FastBayesianContentAdaption. Our approach takes a goal time $t_g$, a design space $\mathcal{D}$, a prior $\mu_0$, a kernel function $k$ and a modified acquisition $\beta$. This algorithm iteratively presents contents $x$ from the design space, measures the interaction with the user $t$, updates $\mu_0$ using a Gaussian Process with kernel $k$ and uses this model to query new contents $x_\text{next}$ that probably have performance close to the goal $t_g$.}
\label{code:FastBayesianContentAdaption}
\end{lstfloat}

\section{Fast Content Adaption through B.O.}\label{sec:FastBayesianContentAdaption}

% TODO: Extend this small presentation. Add a reference to Figure 1.
% \todo{Some more details here. We present a level, player plays it, we update our estimates, ...}

Our approach, called Fast Bayesian Content Adaption (FBCA),
% \todo{change, also in algorithm 1, and figures?}
uses Bayesian Optimization to select the best contents to present to a player in an online fashion. 
%B.O. is a scheme for optimizing functions that have no closed form and must be accessed through expensive queries, rendering it ideal for this application. 
On a high-level, our approach works as follows (Fig.~\ref{fig:banner}): (1) Start a B.O. scheme with a hand-crafted prior over a set of levels/tasks\footnote{Having a handcrafted prior is optional, since the system would also work with a non-informative one.}; (2) present the player an initial guess of what might be a level with the right difficulty and record the player's interaction with it; (3) update our estimates of the player and continue presenting levels that have ideal difficulty according to the internal model.  %Fig.~\ref{fig:banner} shows how this method works for a single player.
% \todo{reference figure 1 in this paragraph}

In normal applications of B.O., the approximated function is precisely the one to be optimized. In our approach, however, we separate the optimization from the modeling using a modified acquisition function. 

\subsection{Modeling variables of interest using Gaussian Processes}

Given a design space $\mathcal{D}$ (e.g. a collection of levels in a video game, or a set of possible web sites with different layouts), it is of interest to model a function $t\colon\mathcal{D}\to\mathbb{R}$. The first step of our approach is to approximate this variable of interest $t(x)$ using Gaussian Process Regression.

In the context of DDA, we choose to model the logarithm of the time $\log(t(x))$ it takes a player to solve task $x$ using Gaussian Process Regression, and we will optimize it for a certain goal time $t_g$ using Bayesian Optimization. We model log-time instead of time to ensure that our estimates of $t(x)$ are always positive. Instead of using time as a proxy for difficulty, other metrics such as score, win rate or kill/death ratio could be chosen, depending on the nature of the game. 

\subsection{Separating optimization from modeling}

Once a model of the player $t(x)$ has been built, we can optimize it towards a certain goal $t_g$. Since the acquisition function in a B.O. scheme is built for finding maximums, we separate modeling from optimization to be able to maximize the target $t_g$: once we regress $t(x)$, we consider $\tilde{t}(x) = - (t(x) - t_g)^2$ as the objective function. After this transformation, the function $\tilde{t}(x)$ has its optima exactly at $t_g$, and the acquisition function is seeking an improvement in $\tilde{t}$ instead of $t$, which translates to values for $t(x)$ that are close to $t_g$.

In other words, we modify the typical acquisition functions (Sec.~\ref{sec:related_work}) in the following way: Expected Improvement becomes $\beta_{\text{EI}}^{t_g} = \mathbb{E}_{\log(t(x))\sim\text{GP}}[\max(0, -(t(x) - t_g)^2 - \tilde{t}_\text{best}]$ (where $\tilde{t}_\text{best}$ is the value of $\{\tilde{t}(x_i)\}_{i=1}^n$ closest to $0$ (i.e. $t\approx t_g$), and the Upper Confidence Bound becomes $\beta_{\text{UCB},\kappa}^{t_g} = -(\exp(\mu(x) + \kappa\sigma(x)) - t_g)^2$ (where $\mu(x)$ and $\sigma(x)$ are the posterior mean and variance according to the Gaussian Process). Since we need to compute expectations (in the case of EI), we perform ancestral sampling and Monte Carlo integration: compute several samples of $\log(t(x))\sim\text{GP}$, and average the argument of the expectation inside $\beta_{\text{EI}}^{t_g}$.

% Notice that we could instead be modeling and optimizing $\tilde{t}$. We \textit{separate} optimization and modeling because are still interested in knowing $t(x)$ for all $x\in\mathcal{D}$, and if we were to model $\tilde{t}$ instead, that information would be lost.
% \todo{and because we think log(t) is normal dist, which neither t nor tilde t is. }

Our approach is summarized in pseudocode in Algorithm \ref{code:FastBayesianContentAdaption}. Once a player starts interacting with our system, we maintain pairs $(x, \log(t))$ and we apply B.O. to serve levels which will likely result in $t(x) \approx t_g$. Notice that maintaining all pairs of tasks and performance, we assume that the player is \emph{not} improving over time. This could be addressed by having a sliding window in which we forget early data that is no longer representative of the player's current skill. In our experiments, however, we update our prior on player's performance with all the playtrace of a given player.

\begin{figure*}[h]
    \centering
    \begin{subfigure}[b]{1\columnwidth}
        \includegraphics[width=\textwidth]{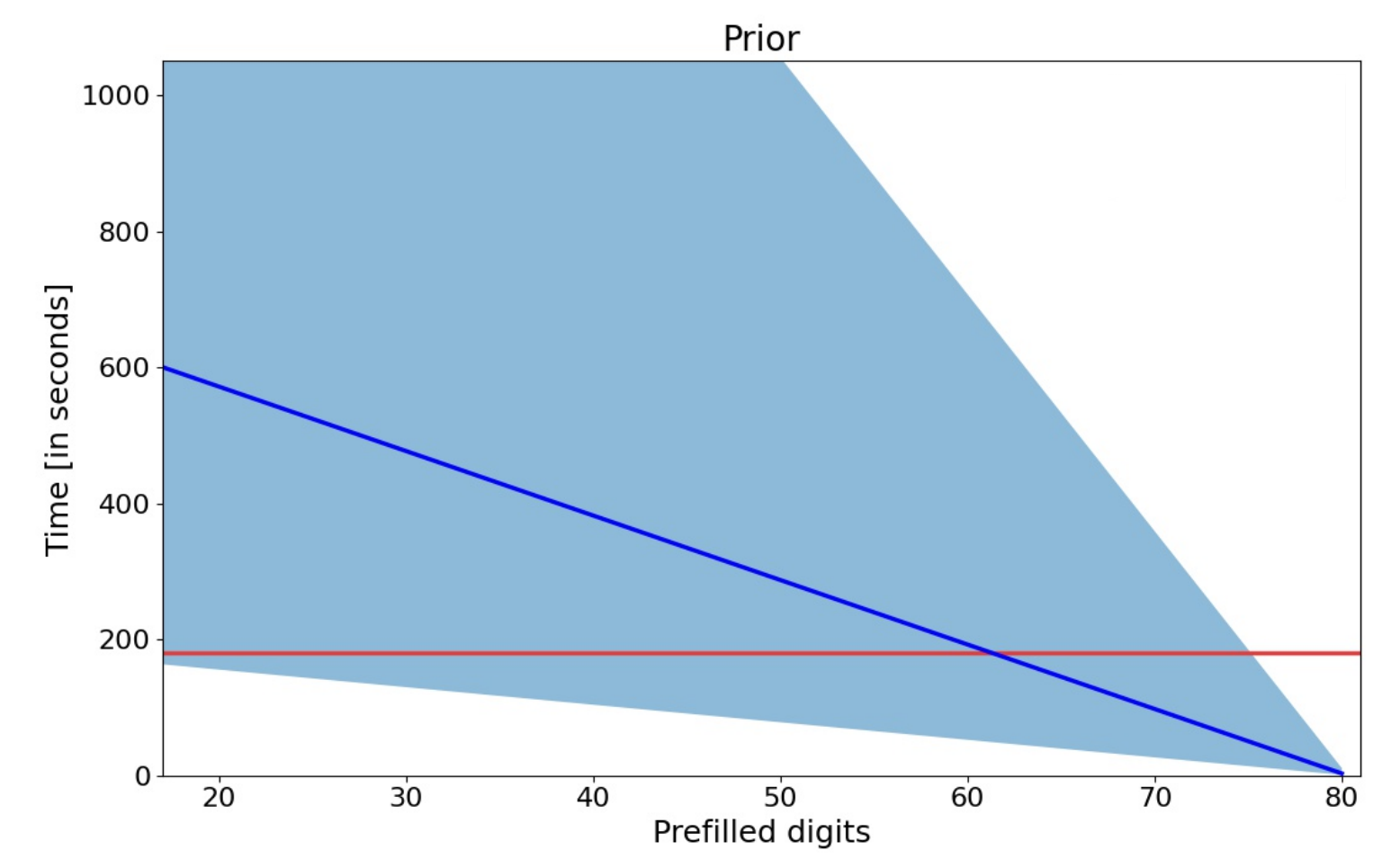}
        \caption{\textbf{Prior for the Sudoku experiment.} This prior encodes the assumption that a Sudoku with 80 prefilled digits would be trivial to solve, and that the difficulty and uncertainty increases as there are fewer prefilled digits.}
        \label{fig:sudoku_prior}
    \end{subfigure}
    \begin{subfigure}[b]{1\columnwidth}
        \includegraphics[width=\textwidth]{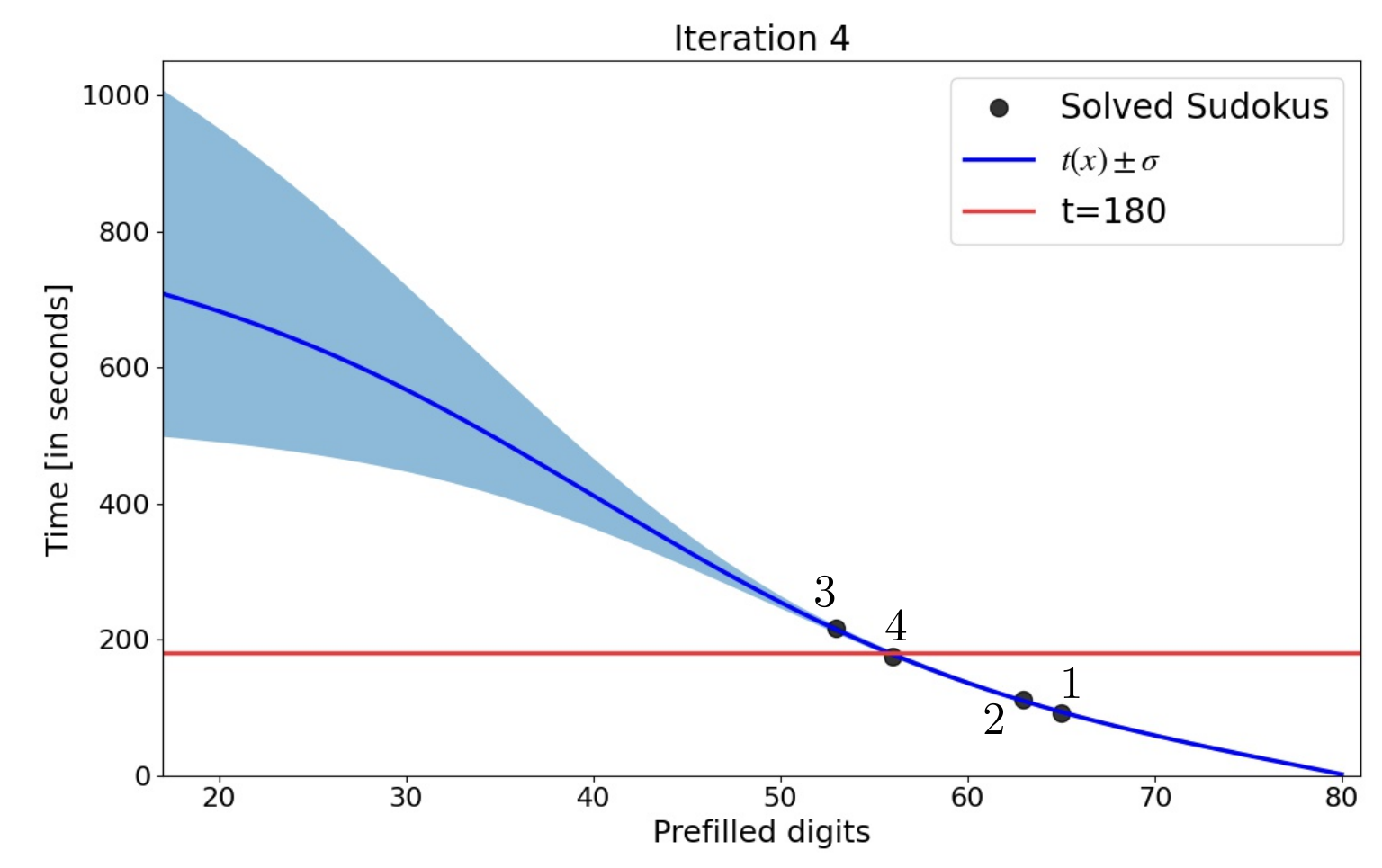}
        \caption{\textbf{4th iteration of a FastBayesianContentAdaption playtrace.}\vspace{0.63cm}} % What a hack.
    \end{subfigure}
    \caption{\textbf{A Sudoku playtrace.}  Once FBCA is deployed with an appropriate prior (a), the player is shown a Sudoku with 65 digits. In the case of this particular playtrace, the player took 92 seconds to complete it. Our approach then runs an iteration of the Bayesian Optimization, serving a level with 63 hints. Since this level is still too easy (111 seconds), the modified acquisition function $\beta^{180}_\text{EI}$ suggests a Sudoku with 53 digits, which takes the player 216 seconds to solve. Finally, a Sudoku with 55 hints is presented that takes the player 175 seconds to solve. The resulting model of the player at this fourth iteration of FBCA is presented in (b).}
    \label{fig:sudoku_prior_and_iterations}
\end{figure*}
% \todo{We don't mention (b) in Figure 2}

\begin{figure}[t]
    \centering
    \includegraphics[width=0.8\columnwidth]{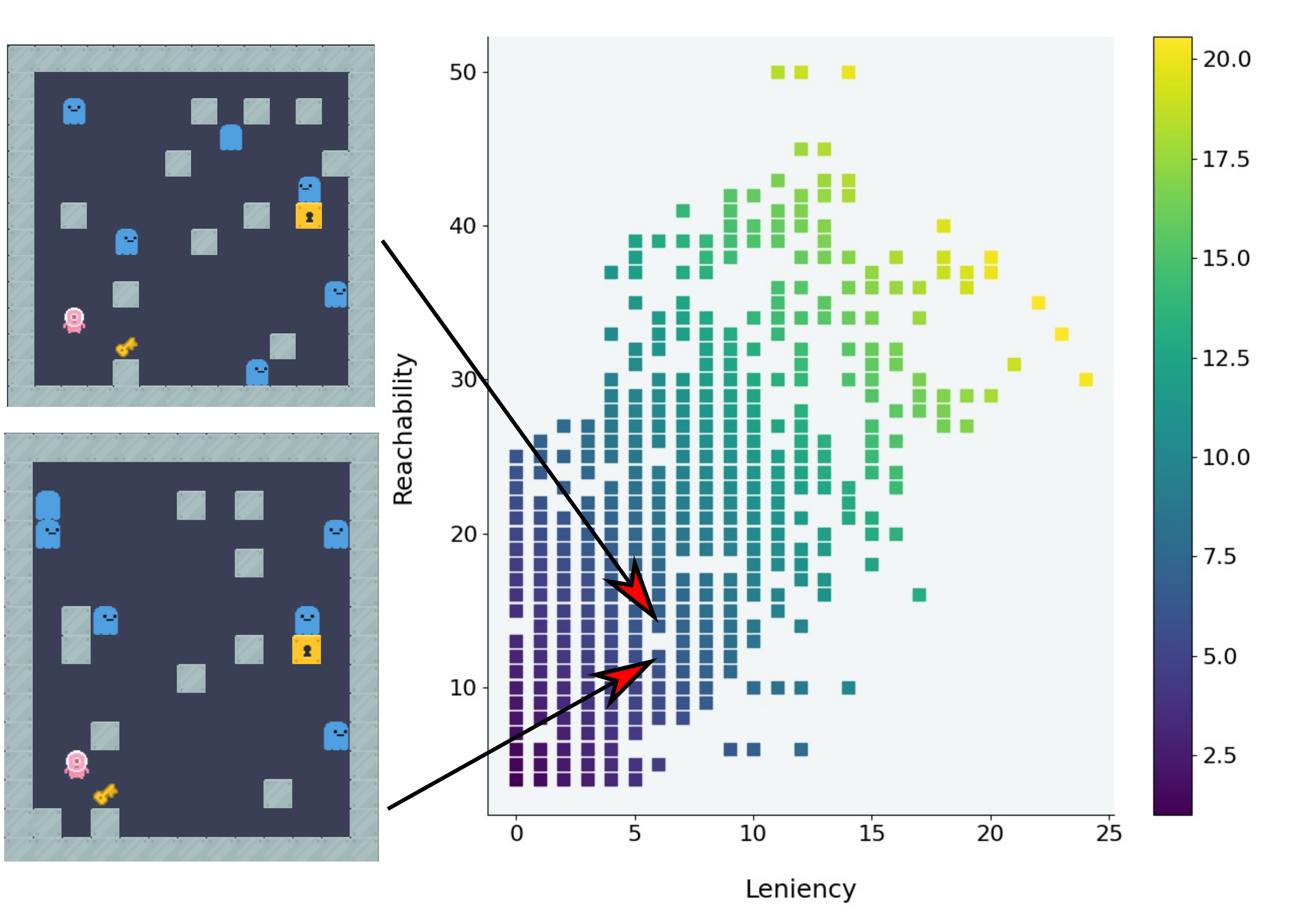}
    \caption{\textbf{Prior creation for the Roguelike experiment.} We pre-computed a corpus of levels for our Roguelike game and, for each one of them, we computed their \textbf{leniency} $l$ (amount of enemies) and \textbf{reachability} $r$ (sum of the length of the A* paths from the avatar to the key and from the key to the final goal). All optimizations of the roguelike experiment take place in the $(l,r)$ space this figure illustrates. We associated to each level a prior, built on the assumption that levels with low leniency and reachability could be solved in 1 second, and levels high leniency and reachability in 20 seconds.}
    \label{fig:prior_creation}
\end{figure}

\section{Experimental Setup}\label{sec:experimental_setup}

% \todo{looks like we referenced the pseudocode, maybe not necessary again here}
Our approach for Fast Bayesian Content Adaption promises to be able to adjust contents in a design space $\mathcal{D}$ according to the behavior of the player. To test this approach we chose two domains: the puzzle Sudoku, and a simple Roguelike game.

% \todo{do we test different priors? reviewer could wonder if we evaluated different priors or what effect they might have. We ran experiment with artificial agents on different priors, right? Maybe we should add some of those results as well. Or maybe we could use the results of the bugged version for comparison (a prior that was artificial made worse).}
% MGD: I would love to include this, but I don't think we have the space. With all the experiments we ran, I think we could easily expand this paper into a journal article.

\subsection{Sudoku}

%  1. Sudoku
%       1. We deployed a website and recorded {...} playtraces.
%       2. Design space: pre-filled digits, sudokus created at random from there.
%       3. We tested two approaches:
%          1. Ours, {something something metadata}.
%          2. Binary search {slight description}.
%       4. Comparison: violin plots for the first {n} iterations, retention funnels, distance-to-goal means...
%       5. Even though the amount of data is quite different for both approaches, it seems that ours performs **on-par** with binary search, with the plus of actually modelling the player's behavior.

A Sudoku puzzle (size 9$\times$9) requieres filling in the digits between 1 and 9 in grids of size 3$\times$3 such that no number appears more than once in its grid, row, and column. The difficulty of a Sudoku can be measured in terms of the number of prefilled digits, which ranges from 17 (resulting in a very difficult puzzle) to 80 (which is trivial). Thus, we settled on $\mathcal{D} = \{17, 18, \dots, 80\}$ as the design space for this domain.

To test our approach, we deployed a web app\footnote{\url{https://adaptivesudoku.herokuapp.com/}} with an interface that serves Sudokus of varying amounts of prefilled digits $x\in\mathcal{D}$. We settled for a goal of $t_g = 180$sec, instructed users to solve the puzzles as fast as they could, and initiated the player model with a linear prior $\mu_0(x)$ that interpolates the points (80, 3) and (17, 600) (see Fig.~\ref{fig:sudoku_prior}). For this experiment, we used the RBF kernel $k_\text{RBF}$ and the modified Expected Improvement acquisition $\beta^{t_g} = \beta^{t_g}_{\text{EI}}$, and we computed this expectation using ancestral sampling (since $\log(t(x))\sim\text{GP}$) and Monte Carlo integration (see Sec. \ref{sec:FastBayesianContentAdaption}). 
% \todo{move acq. fn. max. by ancestral sampling to section III}.
% \todo{move the log(t) explanation to III.A}

We compare our FBCA approach with a binary search baseline which explores $\mathcal{D}$ by halving it, and jumping to the easy/difficult part according to the performance of the player. We also compared our approach with a linear regression policy for selecting the next Sudoku: we approximate player's completion time starting with the same prior as with our approach and, once new data arrives, we fit a linear model to it in log-time space and present a sudoku that the model says will perform closest to $t_g$. Unfortunately, we were not able to gather enough playtraces to make significant statistical comparisons in the linear regression case, so our analysis will focus on the comparison between our approach and simple binary search. However, we present a short comparison between leveraging this linear regression model and deploying our FBCA system.
% \todo{this sounds like we will not show the comparison to the linear model} 

%These sudokus were taken from Kaggle's one-million-sudokus competition\footnote{\url{https://www.kaggle.com/bryanpark/sudoku}}, and a system was developed to allow for serving random sudokus with a specified amount of different amounts of prefilled digits.

% \todo{What dimensions do the maps have in Sudoku and roguelike. why did we decide on Leniency, etc.}

% Describe the design space, the goal and the acq. function (EI with ancestral sampling).

% We tested two different approaches:

%     ours: We used B.O. by modelling the time it takes to a player to solve a sudoku with $x$ prefilled digits with a 1-D GPR. We "bent" the acquisition function to target 3 minutes, and we used so-and-so Kernel.
% FIG: Creating level for the Roguelike

%     a baseline: binary search: starting at {some} amount of hints, we go to the middle point of the interval... 

\subsection{Roguelike}\label{subsec:setup_rogue}

We also tested our approach in a simple Roguelike game, in which the player must navigate a level, grab a key and proceed to a final objective while avoiding enemies that move randomly. The player can move in all directions and when facing an enemy, can kill it (Fig.~\ref{fig:prior_creation}).
%. Example levels of this game can be seen in Fig.~\ref{fig:prior_creation}.

% \todo{I guess we don't know if they are familiar with roguelike games. maybe say "It's an instance of a rogue-like game the players have not played prior}
This domain is significantly different from Sudoku since difficulty is not as straight-forward to define and model, and it is also an instance of a roguelike game that players have not played prior. Two variables that influence the difficulty of a level are \textbf{leniency} $l$, defined as the number of enemies in the level; and \textbf{reachability} $r$, defined as the sum of the A* paths from the player's avatar to the key and from the key to the goal. Using these, we define the as design space $\mathcal{D} = \{(l,r)\}\subseteq\mathbb{R}^2$.

Users are presented levels of the Roguelike game online\footnote{\url{https://adaptive-roguelike.herokuapp.com/}}, and are instructed to solve them as quickly as they can. In our optimizations, we target to find levels that take $t_g=10$ seconds.

We construct a prior $\mu_0(x)$ by computing a corpus of 399 randomly-generated levels $\mathcal{L}$ in the intervals $l\in [0, 24]$ and $r\in [4, 50]$. Levels were generated in an iterative process: first by sampling specifications such as height, width, and amount of enemies and randomly placing assets such that the A* path between the player and the goals are preserved, and then by mutating (adding/removing rows, columns, and assets) previously generated levels. We defined $\mu_0(x)$ by interpolating a plane in which a level with $l=0, r=4$ takes $1$ second to solve, and a level with $l=14,r=50$ would take 20 seconds to solve. Fig.~\ref{fig:prior_creation} shows this prior, as well as example levels and their place in the design space.

For the GP, we chose a kernel $k$ that combines a linear assumption, as well as local interactions using the RBF kernel: $k = k_\text{RBF} + k_\text{Linear}$, together with the modified Upper Confidence Bounds acquisition function $\beta^{t_g} = \beta^{t_g}_{\text{UCB}, \kappa}$ with $\kappa=0.05$. These hyperparameters (kernel shape and $\kappa$) were selected after several trial-and-error iterations of the experiment.
% alongside a grid hyperparameter search using ``artificial agents'' in the form of noisy functions that stand as a proxy for humans.\todo{I don't think we should mention this}
Hyperparameters inside the kernels themselves were obtained by maximizing the likelihood w.r.t the data.

% \todo{we should say why we don't compare to other published methods. Are they not applicable?}
We compare our approach with two other methods: selecting levels completely at random from the corpus $\mathcal{L}$, and a Noisy Hill-climbing algorithm that starts in the center of $\mathcal{D}$, takes a random step (governed by normal noise) arriving at a new level $x$ and performance $t$, and consolidates $x$ as the new center for exploration if $t$ is closer to the goal $t_g$. When users enter the website and game, they are assigned one of these three approaches at random. These baselines were selected since methods for DDA discussed in Sec.~\ref{sec:related_work} are either domain-specific, or they optimize for engagement and thus do not apply in this context.
%\todo{move the log(t) explanation to III.A}

\section{Results and Discussion}

%In this section we discuss the experiments we performed in order to assess how fast our approach (FastBayesianContentAdaption, see Algorithm~\ref{code:FastBayesianContentAdaption}) is capable of adapting to the user. We tested our approach in the domains of Sudoku puzzles, and levels for a simple Roguelike game,  comparing it to simpler baselines in both cases.

\subsection{Sudoku}

% FIG: Fitting a GP with all sudoku playtraces. We don't really need it.
% \begin{figure}
%     \centering
%     \includegraphics[width=1\columnwidth]{sudoku_all.jpg}
%     \caption{All the playtraces for Sudoku}
%     \label{fig:my_label}
% \end{figure}

%Using our prototype web application, players were shown the Sudokus selected by FastBayesianContentAdaption and by binary search, searching for puzzles that would take users 180 seconds to solve. They were instructed to solve these tasks as quickly as they could. 
We received a total of 288 unique playtraces for FBCA, and 94 unique playtraces using binary search. This discrepancy, however, is accounted for in the statistical tests we perform.

% TODO: plot sudokus if I have the time.
%\todo{show some examples generated Sudoku puzzles with different difficulties}

% FIG: A single sudoku playtrace

Figure~\ref{fig:sudoku_prior_and_iterations} shows on the one hand the prior $\mu_0(x)$ we used for the fast adaption, and on the other the approximation $t(x)\sim\text{GP}$ after a player was presented with 4 different Sudokus, following Algorithm~\ref{code:FastBayesianContentAdaption}. The first Sudoku had 65 prefilled digits and was too easy for the player, who solved it in only 92 seconds. In the next iteration, the modified acquisition function $\beta^{180}_\text{EI}$ achieved its maximum over $\mathcal{D} = \{17, \dots, 80\}$ at 63 hints. This Sudoku, however, also proved too easy for the player, who solved it in 111 seconds. The next Sudoku suggested by the system had 55 prefilled digits and took the player over 3 minutes to solve and, finally, at the 4th iteration, the system presented a Sudoku that took the player 175 seconds to solve, showing that our method was able to find (for this particular player) a puzzle with difficulty close to the target $t_g=180$ in only 4 iterations. 

% I would absolutely LOVE to include this plot, but I don't think it fits.
% FIG: GP trained on all playtraces
\begin{figure}
    \centering
    \includegraphics[width=0.8\columnwidth]{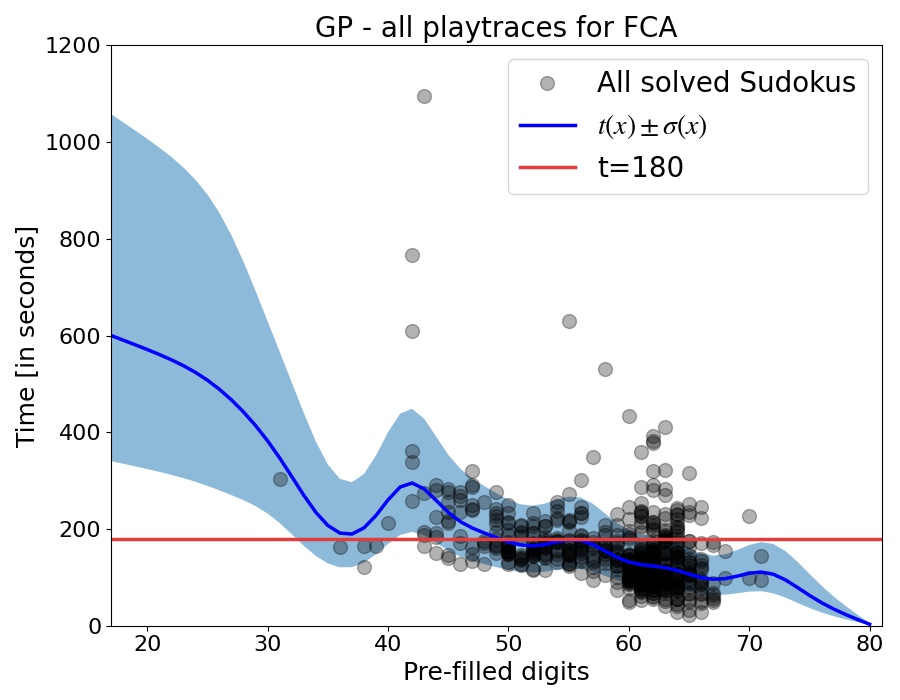}
    \caption{\textbf{The average Sudoku player.} This figure presents the result of fitting a Gaussian process to all the playtraces we gathered for Sudoku when testing the FBCA algorithm. The prior $\mu_0(x)$ presented in Fig.~\ref{fig:sudoku_prior} was updated using an RBF kernel with data from 598 correctly-solved Sudokus, resulting in an approximation $t(x)$ of how long it would take an average player to solve a Sudoku with $x$ hints.}
    \label{fig:sudoku_all}
\end{figure}

% As can be seen when comparing the mean error, there's not a big difference between binary search. Here's a table showing those numbers [ref]

% TABLE: mean errors for the sudokus presented
\begin{table}[]
    \centering
\begin{tabular}{rlll}
\toprule
Iteration & $\mu_e^{\text{FBCA}}\pm\sigma$ & $\mu_e^{\text{bin}}\pm\sigma$ &  $H_0$ rejected \\
\midrule
        1 &       \textbf{72.5}$\pm$ 47.5 &     135.9$\pm$ 266.4 &  yes ($p=0.04$) \\
        2 &       \textbf{76.9}$\pm$ 94.6 &     122.3$\pm$ 202.6 &   no ($p=0.17$) \\
        3 &       \textbf{49.6}$\pm$ 33.8 &     132.7$\pm$ 361.8 &   no ($p=0.22$) \\
        4 &       47.6$\pm$ 66.7 &       \textbf{39.5}$\pm$ 33.2 &   no ($p=0.52$) \\
        5 &      60.4$\pm$ 114.5 &       \textbf{44.1}$\pm$ 35.6 &   no ($p=0.45$) \\
        6 &       \textbf{36.5}$\pm$ 34.9 &      72.6$\pm$ 106.5 &   no ($p=0.35$) \\
        7 &       \textbf{37.3}$\pm$ 36.3 &       49.1$\pm$ 40.1 &   no ($p=0.50$) \\
        8 &       \textbf{29.9}$\pm$ 29.6 &       79.8$\pm$ 79.4 &   no ($p=0.19$) \\
\midrule
      All &      \textbf{63.9} $\pm$ 67.3 &     110.5 $\pm$237.1 &  yes ($p=0.01$) \\
\bottomrule
\end{tabular}
    \caption{\textbf{Mean absolute errors for the Sudoku experiment.} This table presents the mean errors $\mu_e $ and their respective standard deviations. At each iteration, we consider all sudokus solved by users and their respective times. 
    We compute the distance to the goal of the optimization (180 seconds), and we average them over the total amount of sudokus presented for said iteration. 
    The mean error for FBCA $\mu_e^{\text{FBCA}}$ is significantly less than that for the binary search $\mu_e^{\text{bin}}$, which means that the null hypothesis $H_0: \mu_e^{\text{FBCA}} = \mu_e^{\text{bin}}$ is rejected according to a two-tailed $t$-test with acceptance rate of 0.05.
    % This information is also presented visually in Fig.~\ref{fig:mae_sudoku}. NOTE FOR SEBASTIAN: I think we could do without this table. All this information is already being presented in the figure I reference.}
    }
    \label{tab:mae_sudoku}
\end{table}
\begin{table}[t]
    \centering
    \resizebox{\columnwidth}{!}{
\begin{tabular}{llllll}
\toprule
         level \#s & $\mu_e^{\text{FBCA}}$ & $\mu_e^{\text{NH}}$ & $\mu_e^{\text{rand}}$ & $H_0^{\text{NH}}$ rejected & $H_0^{\text{rand}}$ rejected \\
\midrule
   $1 \leq i < 5$ &       \textbf{3.66}$\pm$ 3.48 &      5.76$\pm$ 5.63 &        7.09$\pm$ 6.73 &             yes ($p=0.01$) &               yes ($p<0.01$) \\
  $5 \leq i < 10$ &       \textbf{3.65}$\pm$ 3.94 &      4.88$\pm$ 4.98 &        6.28$\pm$ 6.03 &             no ($p=0.10$) &               yes ($p<0.01$) \\
 $10 \leq i < 15$ &       \textbf{3.97}$\pm$ 5.15 &      4.39$\pm$ 6.76 &        4.65$\pm$ 3.80 &              no ($p=0.72$) &                no ($p=0.47$) \\
 $15 \leq i < 20$ &       4.76$\pm$ 6.62 &      \textbf{3.69}$\pm$ 3.09 &        6.39$\pm$ 6.39 &              no ($p=0.41$) &                no ($p=0.32$) \\
 $20 \leq i < 25$ &       \textbf{3.30}$\pm$ 3.68 &      4.36$\pm$ 4.91 &        5.13$\pm$ 5.45 &              no ($p=0.42$) &                no ($p=0.20$) \\
 $25 \leq i < 30$ &       \textbf{2.23}$\pm$ 1.68 &      5.20$\pm$ 6.17 &        3.97$\pm$ 3.15 &             no ($p=0.07$) &               no ($p=0.08$) \\
 $30 \leq i < 35$ &       \textbf{2.58}$\pm$ 2.44 &      4.00$\pm$ 4.05 &        4.55$\pm$ 1.91 &              no ($p=0.29$) &               no ($p=0.06$) \\
 \midrule
              All &       \textbf{3.78}$\pm$ 4.59 &      4.73$\pm$ 5.32 &        5.94$\pm$ 5.74 &             yes ($p=0.01$) &               yes ($p<0.01$) \\
\bottomrule
\end{tabular}
}
    \caption{\textbf{Mean absolute errors for the Roguelike experiment.} This table presents the mean absolute error for FastBayesianContentAdaption (FBCA) $\mu_e^\text{FBCA}$, the Noisy Hill-climbing (NH) baseline $\mu_e^\text{NH}$, and for serving random levels $\mu_e^\text{rand}$. This data supports two claims: Our approach FBCA improves in its understanding of the player over time (which can be sensed from the fact that $\mu_e^\text{FBCA}$ decreases the more levels a player solves), and FBCA performs better than NH and serving random levels on average (since $\mu_e^\text{FBCA}$ is less than $\mu_e^\text{NH}$ and $\mu_e^\text{rand}$ in almost all intervals). While our approach performs significantly better when taking all iterations into account, in some instances this performance increase, while observable, is not significant. These results are also reported visually in Fig.~\ref{fig:mae_roguelike}}. 
    \label{tab:mae_roguelike}
\end{table}

\begin{figure*}
    \centering
    \begin{subfigure}[b]{1\columnwidth}
        \includegraphics[width=\textwidth]{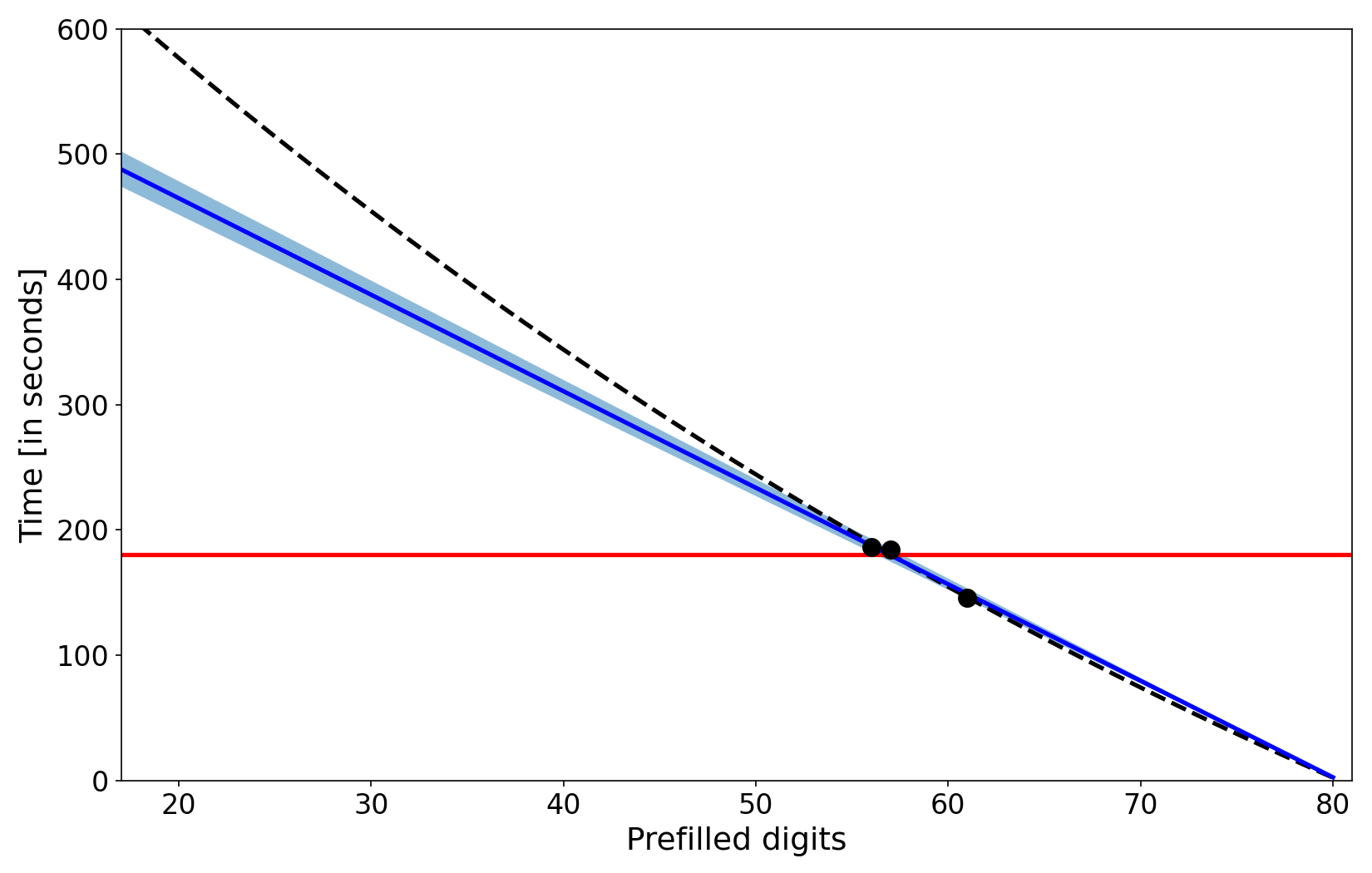}
        \caption{\textbf{Example: linear regression of log-time is enough}.}
        \label{fig:non_problematic}
    \end{subfigure}
    \begin{subfigure}[b]{1\columnwidth}
        \includegraphics[width=\textwidth]{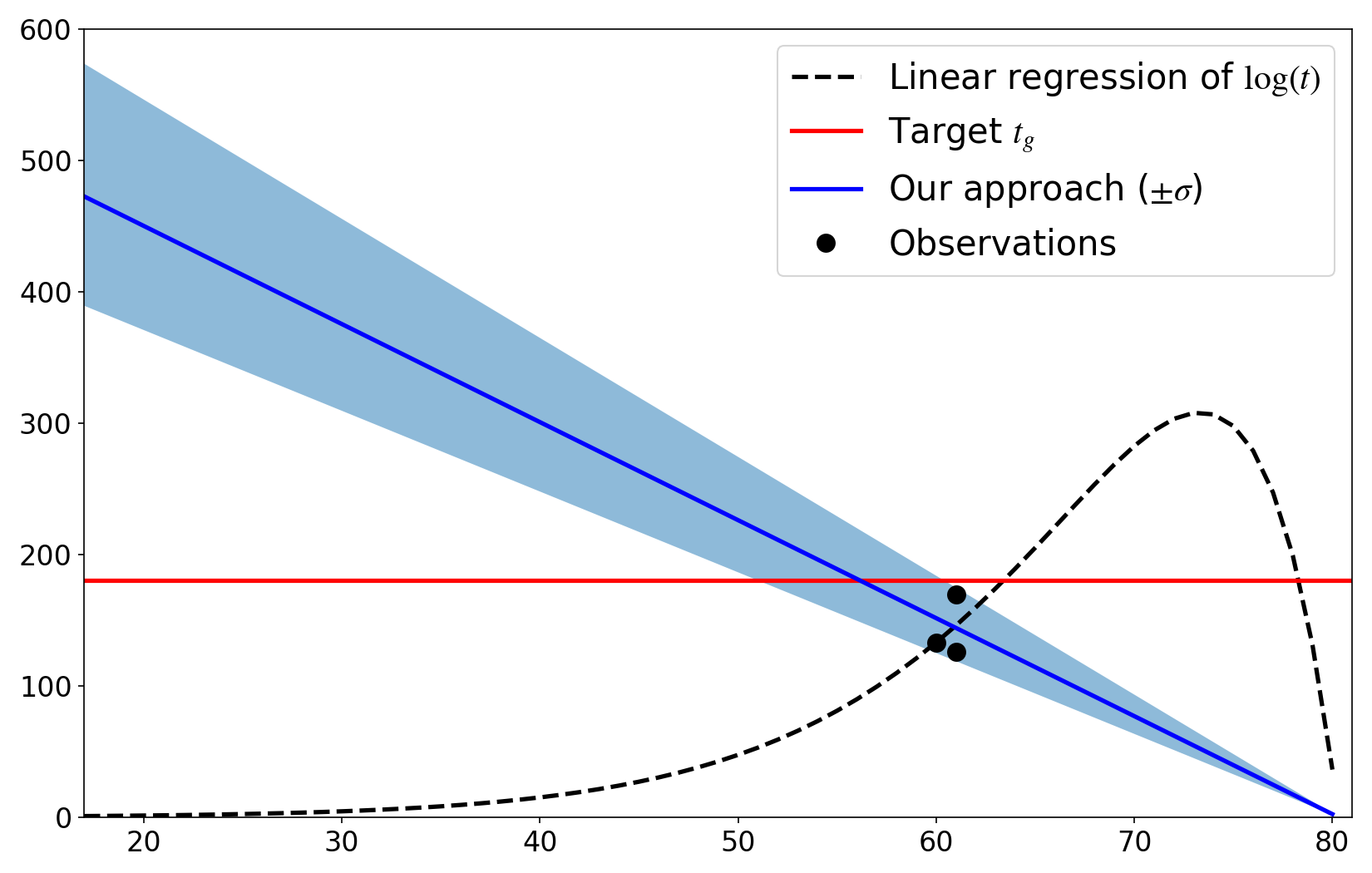}
        \caption{\textbf{Example: linear regression fails to show appropriate sudokus.}} % What a hack.
        \label{fig:problematic}
    \end{subfigure}
    \caption{\textbf{Gaussian Process Regression vs. Linear regression in Sudoku.} Shown is a preliminary comparison of our approach against a greedy policy that models players time using linear regression in $\log(t)$ space. (a) shows an example in which this greedy policy was able to find a sudoku with the target time (with a playtrace of [145, 186, 184] seconds), while in (b) we see an example in which the linear regression fails to capture an adequate model of difficulty for the player (with a playtrace of [169, 132, 125] seconds). In both examples, our approach (that relies on Gaussian Process Regression with an RBF kernel) is able to capture a model of the player that decreases completion time w.r.t increasing hints. This seems to be a sensible model of difficulty, both intuitively and according to the result of fitting our approach with all (solved) sudoku playtraces (see Fig.~\ref{fig:sudoku_all})}
    \label{fig:sudoku_linear_regression_vs_ours}
\end{figure*}

\begin{figure*}
    \centering
    \includegraphics[width=2\columnwidth]{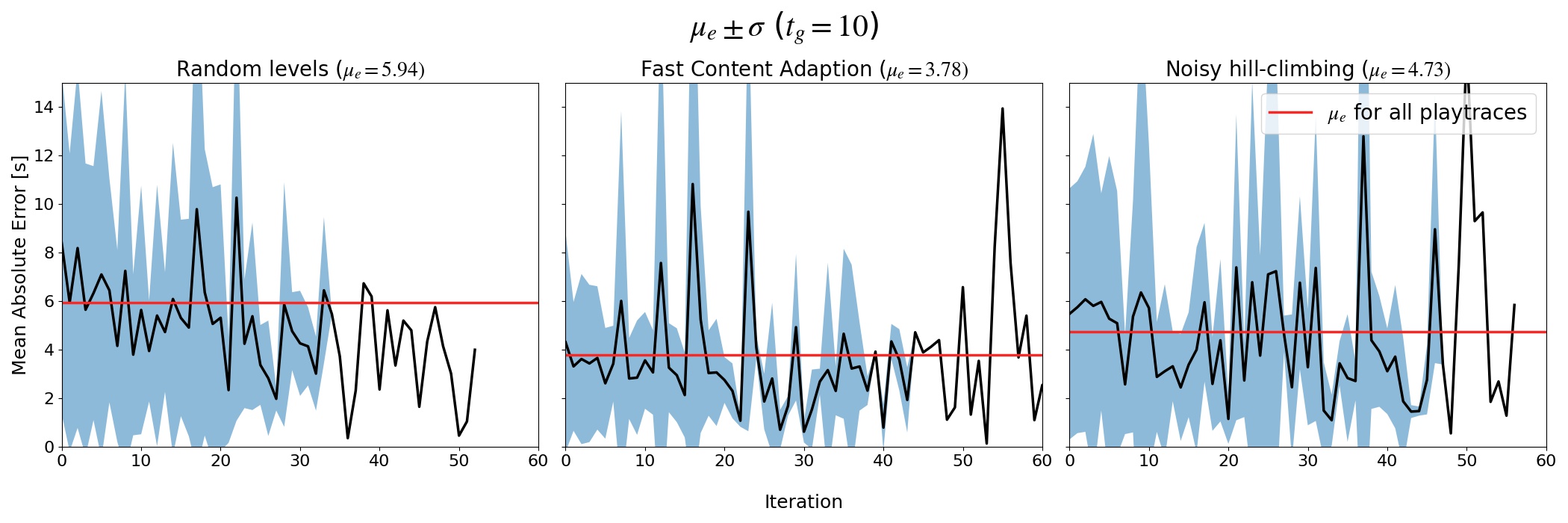}
    \caption{\textbf{Mean absolute error for the Roguelike experiment.} Shown is the average distance to the goal $t_g=10$ for three different approaches (presenting levels randomly, using Fast Bayesian Content Adaption, and using a baseline of Noisy Hill-climbing). Overall, we can see that the Bayesian optimization's absolute mean error is decreasing, albeit with noise. The difference between the Bayesian Optimization and the other approaches is small (in the ballpark of 1 or 2 seconds) but significant. The blue shaded region represents $\pm \sigma$, and it disappears after e.g. 45 in the FBCA since, from that point onwards, we only have one playtrace.}
    \label{fig:mae_roguelike}
\end{figure*}

To measure which approach performed better (between FBCA and binary search), we use the \textbf{mean absolute error} metric, defined as
\begin{equation}\label{eq:mae}
    \mu_e(\mathcal{T};t_g) = \frac{1}{|\mathcal{T}|}\sum_{t\in\mathcal{T}}|t-t_g|,
\end{equation}
this quantity measures how far, on average, the times recorded were from the goal time $t_g$. For brevity in the presentation, we will use the symbols $\mu_e^\text{FBCA}$ to stand for the mean absolute error of the times collected for the Bayesian experiment $\mathcal{T}_\text{FBCA}$ and goal $t_g=180$, and likewise for the binary search data with $\mu_e^\text{bin}$.

Table~\ref{tab:mae_sudoku} shows the mean absolute error for the Sudoku experiment, segmented by the $i$th sudoku presented to the user. We only consider puzzles that were correctly solved, ignoring Sudokus solved in over 3000 seconds. According to the data for the first iteration, Sudokus with 65 hints (the starting point for FastBayesianContentAdaption) are closer to the goal of 3 minutes than the starting point of binary search on average. Both approaches appear to be getting closer to the goal the more Sudokus the player plays, ending with an average distance to the goal of about 30 seconds for FBCA in the best case ($i=8$) and of about 40 seconds in the binary search ($i=4$).
% These different mean absolute errors are displayed in Fig.~\ref{fig:mae_sudoku}, alongside their standard deviations.

% This results must be taken with a grain of salt: high variances do not allow us to reject the null hypothesis $H_0: \mu_e^\text{FBCA} \geq \mu_e^\text{bin}$ (using a one-tailed $t$-test and assuming different group variances, and rejecting $H_0$ only when the $p$-value is below the usual acceptance rate of $0.05$)
While these results have high variance, two-tailed $t$-tests that assume different group variances allow us to reject the null hypothesis $H_0: \mu_e^\text{FBCA} = \mu_e^\text{bin}$ (with a $p$-value of 0.01, less than the acceptance rate of 0.05) when including all the playtraces for both experiments. FBCA has a mean absolute error that almost halves the binary search one (with statistical significance). We hypothesize that this happens because the search in FBCA takes place around the easier levels, as is shown in Fig.~\ref{fig:sudoku_all}.
We assume this preference for easier puzzles is the result  of  a prior that favors easy levels and a variance that is more uncertain of harder levels. The same is not the case for the binary search; if a player solves the first Sudoku in less than 3 minutes, they are presented  with a puzzle with only 33 prefilled digits next (which is significantly harder).

%As we discussed in the experimental setup, we were not able to gather enough playtraces to perform significant statistical comparisons between our approach and using a greedy policy based on linear regression. 
% \todo{sudoku or Sudoku?}
While we were not able to gather enough playtraces to perform significant statistical comparisons between our approach and using a greedy policy based on linear regression,  we can anecdotally tell that there were cases in which  a linear model of log-time failed to capture the player's performance. In Fig.~\ref{fig:sudoku_linear_regression_vs_ours} we show two example playtraces that were gathered for the linear regression policy experiment. In the first one (Fig.~\ref{fig:non_problematic}) we can see that using linear regression was enough to capture a model of the player and accurately present to them Sudokus that targeted the appropriate difficulty. On the other hand, Fig.~\ref{fig:problematic} shows a critical example in which the predictions of simple linear regression state that sudokus with 17 hints are extremely easy. This seems to indicate that Gaussian Process Regression (with RBF kernels) is a better model for predicting completion log-time in this particular experiment.

% FOR SEBASTIAN: I decided to comment this plot for space and because all info is already contained in the table.
% FIG: Error plots for Sudoku
% \begin{figure}
%     \centering
%     \includegraphics[width=1\columnwidth]{mae_error_sudoku.jpg}
%     \caption{\textbf{Mean absolute error for the sudoku experiment.} This plot shows $\mu_e = \frac{1}{n_k}\sum_{t\in S_k} |t - 180|$. This value represents, for each iteration $k$ of the optimization, the average distance to the goal. A bird's eye view shows that our approach (optimizing $|t-180|$ using Bayesian Optimization) achieves a better performance on average when compared to a binary search scheme. However, the shaded areas (which represents $\pm \sigma_e$) show that these results have high variance, resulting in non-significant statistics (see Table~\ref{tab:mae_sudoku}). NOTE FOR SEBASTIAN: I haven't settled for the colors, and I accept suggestions!}
%     \label{fig:mae_sudoku}
% \end{figure}

\subsection{Roguelike}

% \todo{we should introduce that acronym once and earlier}
% \todo{we repeat some things here we already talk about in Section 4. Can remove this or merge}
%The second domain in which we tested FastContentAdaptation (FBCA) consisted of a simple Roguelike game in which players navigate towards a goal while avoiding enemies. 
Players were assigned one of the three experiments discussed in Sec.~\ref{subsec:setup_rogue} at random when using our web application and were instructed to solve the levels they were presented as quickly as possible. We recorded a total of 18 unique playtraces for FBCA, 21 for the Noisy Hill-climbing, and 30 for the completely random approach. Like in the Sudoku experiment, our statistical tests are resilient to this discrepancy in sample sizes. In preparation for the data analysis, we removed all levels solved in over 60 seconds of time, since they are designed to be short, and a solved time of over a minute may indicate that the player got distracted.

% \todo{what is the p value? do we say which test we used?}
Table~\ref{tab:mae_roguelike} shows the mean absolute error (see Eq. (\ref{eq:mae})) for all three approaches, dividing the iterations in smaller intervals for ease of presentation and smoothing purposes. We see that the mean absolute error for FBCA ($\mu_e^\text{FBCA})$ decreases the more levels a player solves. This effect is to be expected since the underlying player model better approximates the player's actual performance. Although the results have high variance, we can say with statistical confidence (using a two-tailed t-test with an acceptance rate of $0.05$) that FBCA performs better than Noisy Hill-climbing and serving completely random levels, but this is due to a small margin: of less than a second vs. NH, and two seconds vs. serving random levels. Fig.~\ref{fig:mae_roguelike} shows the mean absolute error for all three methods, including one standard deviation and the average for all levels solved. Notice how the mean absolute error for all playtraces is lower for FBCA than for serving random levels, or Noisy Hill-climbing. We argue that the position of this mean error also indicates that the high spikes that can be seen in all approaches are outliers that could be explained by temporary distractions from the players. 

% \todo{Discuss broader impact}
\section{Conclusions and Future Work}

%Future work and drawbacks/limitations (dealing with noisy humans, humans get better with practice -> what does that mean for the system)

%\todo{it should be move than an application. e.g. "A new approach for modelling players with ..."}
This paper presented a new approach, called Fast Bayesian Content Adaption (FBCA), with a focus on automatic difficulty adjustment in games. Our method maintains a simple model of the player, updates it as soon as data about the interaction between the user and the game arrives using Gaussian Processes, and uses a modified acquisition function to identify the next best level to present.

% \todo{significantly less?}
We tested FBCA in two domains: finding a Sudoku that takes 3 minutes to solve, and a level for a Roguelike game that takes 10 seconds to solve.  Experiments that compare FBCA with simpler baselines in both domains show that our approach is able to find levels with target difficulty quickly and with significantly less distance to the target goal. While the results have a high variance, the mean error of FBCA is significantly lower than that of the baseline we compare to.
%Since these experiments were performed in uncontrolled environments (using web applications open to the public), these results unfortunately have high variance. However, one-tailed $t$-tests state that the mean error for FBCA is lower than that of the baselines we compare to.

Our method thus provides an alternative for automatic difficulty adjustment that can, in only a few trials, learn which content in a design space elicits a particular target value (e.g. completion time) for a given player. 

There are several avenues for future research: First, enforcing stronger assumptions about the monotonicity of the function that is being modeled in the Bayesian Optimization \cite{Riihimaki2010} might be beneficial if the content of the game is encoded in features that relate to difficulty directly (e.g.\ amount of enemies). Moreover, approaches in which the prior is automatically learned using artificial agents as proxies for humans could be explored \cite{Jeppe2020}. The fact that our model assumes that the player does not improve over time could also be tackled in future research, by \textit{forgetting} the initial parts of the playtrace using a sliding window. Finally, the features that make the contents of the design space difficult could be automatically learned by training generative methods (e.g.\ GANs or VAEs) and exploring the latent space of content they define \cite{marioGAN2018,BootstrappingGANs2020,fontaine2020illuminating,sarkar2020conditional}.

% We adapted the content of a game using a modified scheme of B.O. in two games.
% We built a sudoku web app, and the B.O. approach was as good as binary search.
% We then built a web app for a simple roguelike game, and the results were noisy, but better than a simple hillclimbing algorithm. We compare with pure randomness and got better results, but it seems that other approaches can get to similar performance, albeit with more variance (?).

% There's plenty to do in the future: better encodings for the levels (including automatic encodings), encoding monotonicity in the kernel, trying few-shot B.O, or safer B.O. Also, other distributions are very relevant and would better capture the model.\todo{what about other domains? more complex games? going beyond games and game for cognitive rehabilitation}

% Highlight the novelty w.r.t. previous work.

% 5. Conclusions
%   1. We tested the use of GPs and BO for DDA.
%   2. We ran two experiments that represent games of different types.
%   3. Results say that we perform on par for Sudoku, and {...} for Zelda.
%   4. Future research includes
%       1. Automatic encoding of the levels (using generative models).
%       2. Forcing monotonicity in the kernels.
%       3. Trying few-shot B.O. techniques.
%       4. Using other distributions (e.g. modeling with negative-binomial).
%.      5. Getting better over time.

\section*{Acknowledgements}
This work was funded by a Google Faculty Award 2019, Sapere Aude DFF grant (9063-00046B),   the Danish Ministry of Education and Science, Digital Pilot Hub and Skylab Digital. We would also like to thank the testers of our two prototypes, and the reviewers for their helpful comments on our manuscript.
\bibliographystyle{IEEEtran}
\bibliography{biblio.bib}

\end{document}